\documentclass{article} 
\usepackage{iclr2024_conference,times}

\usepackage{amsmath,amsfonts,bm}

\def\eqref#1{equation~\ref{#1}}

\def\1{\bm{1}}

\DeclareMathAlphabet{\mathsfit}{\encodingdefault}{\sfdefault}{m}{sl}
\SetMathAlphabet{\mathsfit}{bold}{\encodingdefault}{\sfdefault}{bx}{n}

\usepackage{hyperref}
\usepackage{url}
\usepackage{listings}
\usepackage{graphicx}

\usepackage{color}
\usepackage{framed}

\definecolor{shadecolor}{gray}{0.95}
\usepackage{dirtree}
\hypersetup{colorlinks,linkcolor={blue},citecolor={blue},urlcolor={blue}}  

\title{Global Mapping of Exposure and\\Physical Vulnerability Dynamics \\in Least Developed Countries using\\Remote Sensing and Machine Learning}  

\author{Joshua Dimasaka \\
UKRI CDT in AI for Environmental Risks\\
Centre for Risk in the Built Environment \\
Department of Architecture,\\
University of Cambridge\\
Cambridge, United Kingdom\\
\texttt{jtd33@cam.ac.uk} \\
\And
Emily So \\
Centre for Risk in the Built Environment \\
Department of Architecture,\\
University of Cambridge\\
Cambridge, United Kingdom\\
\texttt{ekms2@cam.ac.uk} \\
\And
Christian Geiß \\
German Aerospace Center (DLR) \\
Institute of Geography,\\
University of Bonn \\
Bonn, Germany\\
\texttt{christian.geiss@dlr.de}
}

\iclrfinalcopy
\begin{document}

\maketitle

\begin{abstract}
As the world marked the midterm of the Sendai Framework for Disaster Risk Reduction 2015-2030, many countries are still struggling to monitor their climate and disaster risk because of the expensive large-scale survey of the distribution of exposure and physical vulnerability and, hence, are not on track in reducing risks amidst the intensifying effects of climate change. We present an ongoing effort in mapping this vital information using machine learning and time-series remote sensing from publicly available Sentinel-1 SAR GRD and Sentinel-2 Harmonized MSI. We introduce the development of ``OpenSendaiBench’’ consisting of 47 countries wherein most are least developed (LDCs), trained ResNet-50 deep learning models, and demonstrated the region of Dhaka, Bangladesh by mapping the distribution of its informal constructions. As a pioneering effort in auditing global disaster risk over time, this paper aims to advance the area of large-scale risk quantification in informing our collective long-term efforts in reducing climate and disaster risk.

\end{abstract}

\section{Introduction}

A global concern on the increasing frequency and intensity of climate disasters, the exacerbating effects of climate change, and the higher rate of increase of exposed human settlements despite a decrease in their vulnerability urged the international community to jointly develop the Sendai Framework for Disaster Risk Reduction (SFDRR) 2015-2030 \citep{unisdr2015}. However, in its 2023 midterm review, the United Nations reported that "a lack of quality, interoperable, or accessible data" to quantify \emph{risk} as a product of \emph{hazard}, \emph{exposure}, and \emph{vulnerability} remains a challenge, especially in many least developed countries (LDCs) where data-collection tools have become inequitably unaffordable \citep{undrr2023}. In particular, the expensive large-scale operation to standardize \emph{exposure} datasets (e.g., human settlements) across countries with different and incomplete \emph{physical vulnerability} characteristics (e.g., building material and construction type) has remained the primary bottleneck to providing a reliable understanding and audit of the evolving climate and disaster risk landscape globally \citep{so2023data}.

Early efforts in developing large-scale exposure datasets were able to map the distribution of human settlements and their physical vulnerabilities \citep{gamba2012ged4gem, huyck2019meteor}, which have been the basis of several global assessment reports \citep{undrr2013, undrr2015, undrr2019, undrr2022}. Unfortunately, these datasets contain limited generalizability and inherent biases that favor developed countries. Specifically, LDCs have different and non-standard vulnerability characteristics because of the ubiquity of informal settlements and different construction methodologies \citep{gunasekera2015developing, silva2022building} and are increasingly outdated because of rapid urbanization \citep{so2023data}.

Furthermore, the theme of most data-driven efforts \citep{esch2022world, sirko2021continental} focuses on mere detection of buildings (i.e., a simple binary task to estimate the presence or absence of a building as a geometry feature or a land use class that is inferred from satellite imagery). Despite several geospatial dasymetric efforts using digital elevation (DEM) and surface (DSM) models as a proxy to understand the downscaled distribution of physical vulnerability characteristics \citep{geiss2023benefits}, it remains difficult because of the high acquisition costs vis-à-vis high-resolution quality and the limited temporal availability of DEMs and DSMs globally. Hence, many recent efforts attempted to use indirect datasets such as the publicly available imagery from Sentinel-1 and Sentinel-2 satellites because of its ability to capture the optical and backscattering signatures of the built environment that could be related to pertinent characteristics such as building height, surface color, and roof roughness \citep{muller2023deep, frantz2021national}.

\begin{figure}[!htpb]
    \centering
    \includegraphics[width=\textwidth]{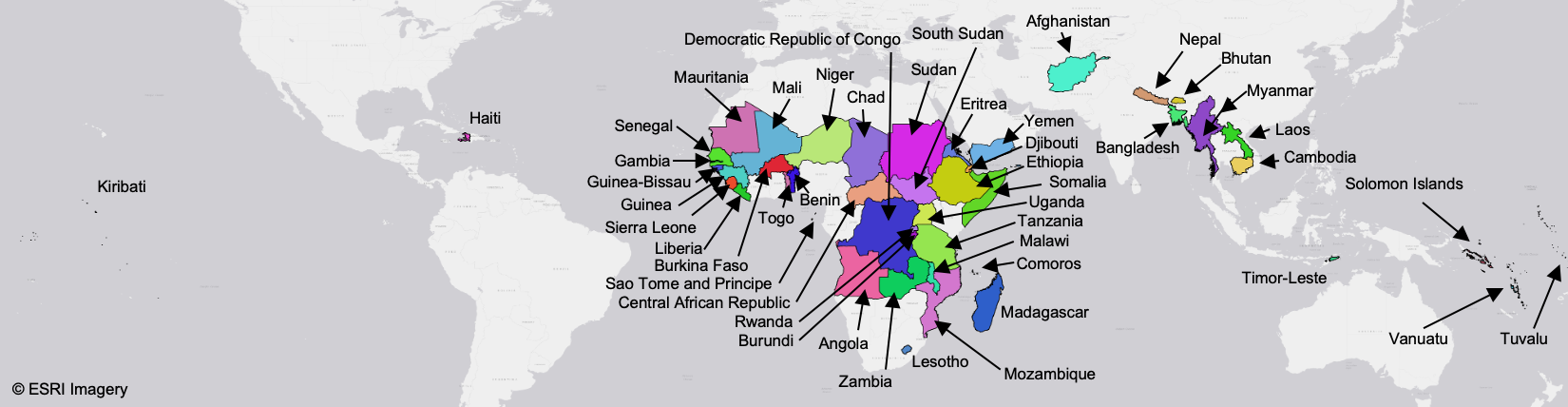}
    \caption{Geographical coverage of the ``OpenSendaiBench'' dataset with 47 countries.}
    \label{fig:country}
\end{figure}

We present an ongoing effort to globally map not only the \emph{exposure} but also its associated \emph{physical vulnerability} characteristics using time-series medium-resolution satellite imagery (i.e., 5-30 meters/pixel). We introduce an initially developed benchmark dataset ``OpenSendaiBench'' (see Figure \ref{fig:country}) and present our findings from a multi-pixel and multi-resolution implementation using the ResNet-50 deep convolutional neural network (CNN) architecture \citep{he2016deep}. Our primary purpose is to bring into awareness this timely and relevant interdisciplinary problem to advance the area of large-scale risk quantification in informing our collective SFDRR and post-2030 long-term efforts.

\section{The ``OpenSendaiBench'' dataset}

The global dataset is a 60-GB collection of 47 countries, wherein 45 are LDCs, and is available in our public Zenodo repository \citep{openSendaiBench} with the following folder structure.\\

\dirtree{%
.1 \textbf{extent}.
.2 [countryCode]\_[nth]\_of\_[totalTiles]\_[index].geojson.
.1 \textbf{groundtruth}.
.2 [countryCode]\_nbldg\_[vulnerabilityCode]\_[nth]*.tif.
.1 \textbf{obsvariables}.
.2 \textbf{SENTINEL1-DUAL\_POL\_GRD\_HIGH\_RES}.
.3 [countryCode]\_[nth]\_of\_[totalTiles]\_[index].
.4 [year]\_VV.tif.
.4 [year]\_VH.tif.
.2 \textbf{SENTINEL-2-MSI\_LVL2A}.
.3 [countryCode]\_[nth]\_of\_[totalTiles]\_[index].
.4 [year]\_RGB.tif.
}

\subsection{National Census-derived Exposure Data}
\label{groundtruth}

We rasterized every country-wide point dataset of building counts from the METEOR project with a defined physical vulnerability type (see \ref{appendix_buildingtypes} for a complete list of typology) at a spatial resolution of 15 arcseconds or approximately 500 meters at the equator \citep{huyck2019meteor}. We then implemented a rigorous probability-based approach in extracting 100 square tiles for each country. We consider the effect of relative areal extent differences of these countries for future work.

In sampling these 100 square tiles, we considered the number of physical vulnerability types that are present in every pixel to ensure that every label including those unlabeled pixels is represented. Specifically, we analyzed the empirical probability distribution of each present physical vulnerability type and assigned a joint probability for each point on the map, assuming that the probability of being in a particular type is independent of each other (e.g., probability of a point being an informal settlement is not influenced by the probabilities of any other building types). We consider the effects of other highly vulnerable types on the agglomeration of informal settlements for future work.

We used these resulting joint probability values as inputs to our importance sampling technique to ensure a balanced representation of sampled pixels. Instead of individual sampling of pixel locations, we extracted square tiles, which is an 8-pixel-by-8-pixel group of 64 sampled locations.

\subsection{Time-series Satellite Imagery}

With the previously extracted geographical extents, we obtained the following pre-processed time-series satellite imagery via Google Earth Engine \citep{gorelick2017google}. 

\textbf{Sentinel-1 SAR GRD.} At 10-m spatial resolution, we used the annual mean of the Ground Range Detected (GRD) scenes that are acquired from the dual-polarization C-band Synthetic Aperture Radar (SAR) instrument at 5.405GHz of Sentinel-1 satellite \citep{sentinel1}. As a result, covering the years from 2019 to 2023, we extracted nine annual mean of the two bands: VV (vertical transmit, vertical receive) and VH (vertical transmit, horizontal receive) signals. To avoid data incompleteness across large areas, we disregarded filtering by orbital number and satellite direction. We also note that there are countries such as Angola, Comoros, Ethiopia, Kiribati, and Tuvalu with either partially or fully complete VV and VH signals because the orbit of Sentinel-1 satellite does not cover these areas for some time or only a single VV signal is available (see \ref{appendix_completness}).

\textbf{Sentinel-2 Harmonized MSI.} With similar spatial resolution at 10 meters, we also extracted the annual median of the atmospherically corrected surface reflectance signals represented by the red, green, and blue (RGB) bands that are acquired from the MultiSpectral Instrument (MSI) of Sentinel-2 satellite \citep{sentinel2}. The aggregation by year also enables minimizing the unnecessary noisy cloudy or shadowy signals using the available and corresponding Sentinel-2 cloud probability dataset \citep{sentinel2cloud}. Unlike Sentinel-1 SAR GRD, the resulting five annual median maps from 2019 to 2023 are all available for 47 countries.

\section{Problem Definition: A Multi-resolution Multi-pixel Framing}

Because of the differing spatial resolutions of ground truth labels and satellite imagery inputs, we approach this as a multi-resolution multi-pixel problem. We hypothesize that the 50x finer resolution of satellite imagery contains significant and detailed spatial patterns that could be informative to the learning of our machine learning models. Hence, we performed the upscaling or aggregation within the ResNet-50 architecture so that the resulting predictions have a similar dimensionality as the ground truth labels. Despite the interesting simpler opportunity to investigate a single-pixel approach, we also assume that a multi-pixel representation in the form of 8x8-array considers the pertinent information from neighboring pixels.

Moreover, we frame the problem to have coarser resulting predictions wherein the ground truth labels are not instead downscaled to match the resolution of satellite imagery because of the ethical consideration where overly localized attribution or prediction may pose social harm and cause undesirable impacts to the applied area of climate and disaster risk, particularly in formulating regional policies in climate financing or insurance. In other words, we note the lower-resolution approach is suitable for not only efficiently conducting large-scale risk quantification but also preserving the privacy of confidential household information as to what kind of building material, physical vulnerability, relative economic valuation, or any other indirect variables that could be inferred from a high-resolution study at 10 meters or finer.

Thus, we define our problem that, in every square tile ${T_{i}}^{year}$ for $i \in \left [ 1, 100 \right ]$, there is a pair of ($x_{locationIndexX}^{signalBand}$, $y_{locationIndexY}^{vulnerabilityType}$) wherein the $locationIndexX$ and  $locationIndexY$ have geographical alignment mapping relationship. For $x$, the $signalBand$ is any combination from the set $\left \{ red, green, blue, VV, VH \right \}$. For $y$, the $vulnerabilityType$ is any combination of physical vulnerability types common to a subset of countries. For example, the ``informal constructions'' type is present in 33 out of 47 countries.

As $y$ values could take extremely high and low values (e.g., a highly dense area with hundreds of buildings), instead of $y$ as a direct input to our CNN model, we computed the $P_{nonexceedance}$ from a lognormal fit of building counts for a particular $vulnerabilityType$, as a random variable (i.e., $y \sim \ln \mathcal{N}(\mu, \sigma^2) \;$). This transformed representation enables the use of probability values limited to the range $\left [ 0, 1 \right ]$, which effectively provides a regional measure of dispersion as a decision variable of interest in the practice of performance-based engineering involving earthquakes and other natural hazards \citep{heresi2023rpbee}. The $P_{nonexceedance}$ is interpreted as the probability that a building count of a particular vulnerability type will be less than the predicted building count, which is computed as:
\begin{equation} \label{nonexceedance}
    P_{nonexceedance} = P\left [ Y_{locationIndexY}^{vulnerabilityType} \leq y \right ] = \Phi(\frac{\ln(y)}{\sigma}) = \frac{1}{2} \left[ 1 + \mathrm{erf}\left( \frac{\ln y-\mu}{\sqrt{2} \sigma} \right) \right]
\end{equation} 

\section{Baseline Experiment: A ResNet-50 CNN Implementation}

Modifying the ResNet-50 CNN architecture for a custom number of input signals, we trained the models with Adam optimizer, an initial learning rate of 0.0001, a batch size of 64, and a training-validation-testing split ratio of 60-20-20. As shown in Table \ref{table}, the numerical findings revealed that the model \textbf{S1} trained with VV and VH signals resulted in the smallest MSE and MAE scores, implying that it can accurately estimate $P_{nonexceedance}$ with an average absolute error of around $\pm$1\%. In addition, the model \textbf{S1} improved the MSE and MAE scores of the model \textbf{S2} by 25\% (MSE) and 22\% (MAE) and of the model \textbf{S1+S2} by 4.1\% (MSE) and 10\% (MAE), respectively. This implies that the backscattering SAR signals, which primarily capture the surface roughness and texture of the ground, were more effective in learning the features than the optical RGB signals.

\begin{table}[!htbp]
    \centering
    \caption{Baseline test set score results for the 'informal constructions' type.}
    \vspace{0.2cm}
    \begin{tabular}{ccc}
    \hline
        \textbf{Model (Input Bands)} & $\textbf{MSE} \left [ 10^{-3} \right ]$ & $\textbf{MAE} \left [ 10^{-2} \right ]$\\ \hline
        \textbf{S1} (VV, VH) & \textbf{4.93} & \textbf{1.07} \\ 
        \textbf{S2} (R, G, B) & 6.56 & 1.38 \\ 
        \textbf{S1+S2} (VV, VH, R, G, B) & 5.14 & 1.19 \\ \hline
    \end{tabular}
    \label{table}
\end{table}

However, we note that additional preprocessing investigation may be needed to meaningfully use the optical RGB signals because these capture the optical signatures such as the roof color of the building. Existing building datasets and elevation maps as a prior belief may also be able to prune and enhance the model capability because other land features such as vegetation that rapidly changes through time may have affected the learned model parameters.

\begin{figure}[!htbp]
    \centering
    \includegraphics[width=\textwidth]{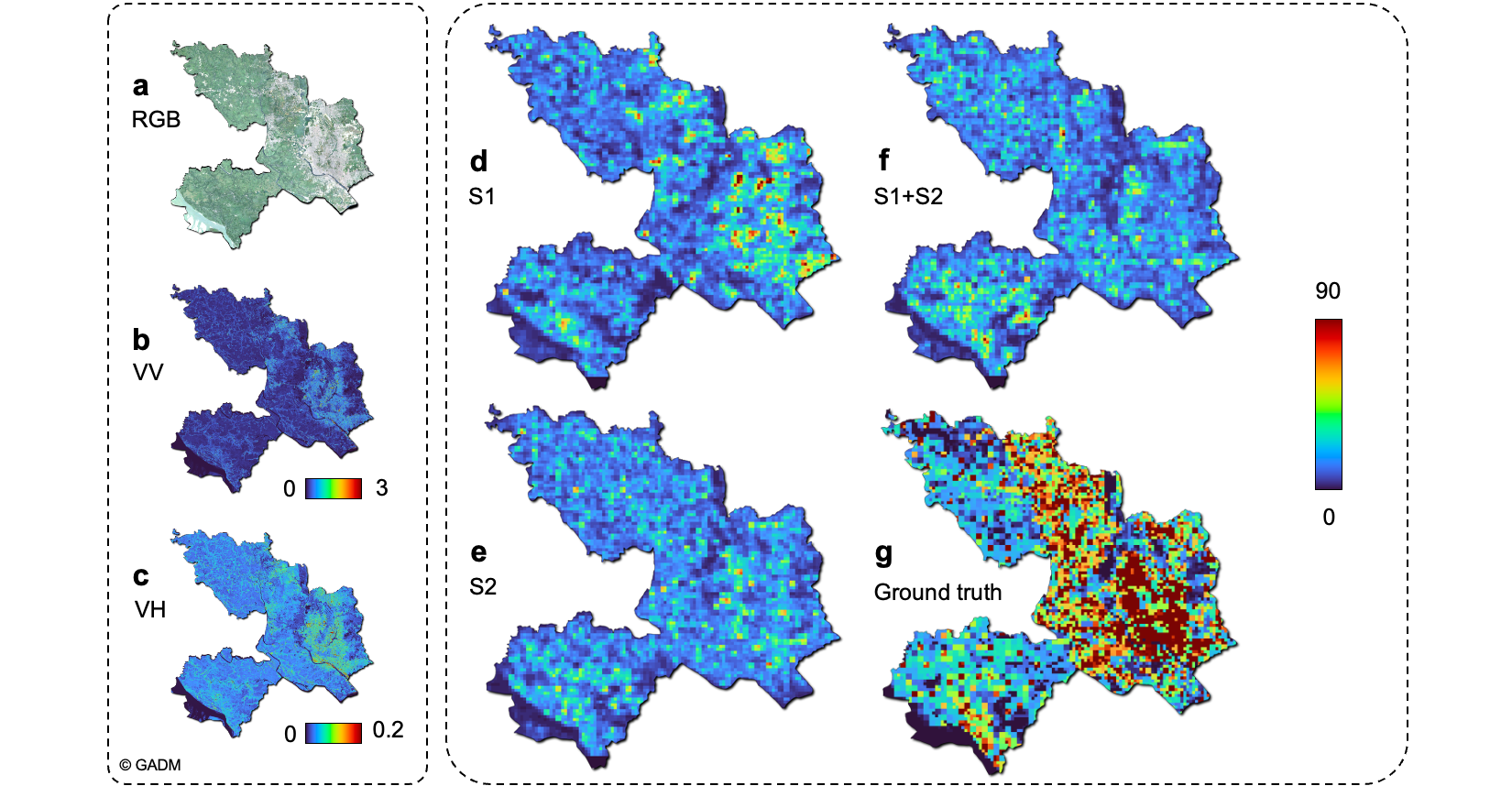}
    \caption{Predicted 2019 distribution of ``informal constructions'' of Dhaka, Bangladesh.}
    \label{fig:dhaka}
\end{figure}

Furthermore, Figure \ref{fig:dhaka} shows the predicted distribution of exposure and physical vulnerability of the city of Dhaka, which has many informal constructions, for the year 2019. We observed that, despite the underestimation of the large values of building count, the models can distinguish the areas with relatively high and low counts, which indicates that the probabilistic transformation should be chosen reliably to represent regional building counts with consideration of extreme values.

\section{Conclusion and Future Work} 

As we are faced with global uncertainty about whether our local and collective efforts in disaster risk reduction have been progressing, we presented an ongoing machine learning effort that addresses this pressing problem to quantify large-scale risk by mapping the exposure and physical vulnerability characteristics using remote sensing. As a pioneering effort, we introduced the development of ``OpenSendaiBench'', a global benchmark dataset that enables the community of both machine learning specialists and disaster risk modelers to contribute and build methodologies. We demonstrated the technical feasibility of this effort using a simple deep-learning model with promising baseline test score results and highlighted the story of informal constructions in Dhaka, Bangladesh.

For future work, we aim to expand the global catalog with Landsat and other Sentinel-2 imagery bands, use elevation maps (DEM/DSM) as prior belief, and incorporate spatial urban morphology growth models to empirically describe the regional dynamics. In partnership with key stakeholders, we plan to localize this effort for some selected cities in the Philippines and Bangladesh. Towards the end, we will implement the probabilistic risk analysis and derive the regional risk metrics, depending on the dominant natural hazards in a given country.

\section{Acknowledgments}
This work is funded by the UKRI Centre for Doctoral Training in Application of Artificial Intelligence to the study of Environmental Risks (EP/S022961/1).

\bibliography{iclr2024_conference}
\bibliographystyle{iclr2024_conference}

\appendix
\section{Appendix}

\renewcommand{\thetable}{\Alph{section}\arabic{table}}
\setcounter{table}{0}

\subsection{List of Building Types}
\label{appendix_buildingtypes}

\begin{table}[!htpb]
\centering
\caption{Description of physical vulnerability types and the number of countries with these types.}
\vspace{0.2cm}
\begin{tabular}{c|c|l} 
\hline
Symbol & Countries & Description                                                                           \\ 
\hline
A      & 32        & Adobe blocks (unbaked sundried mud block) walls                                       \\ 
\hline
C      & 7         & Reinforced concrete                                                                   \\ 
\hline
C3L    & 43        & Nonductile reinforced concrete frame with masonry infill walls low-rise               \\ 
\hline
C3M    & 19        & Nonductile reinforced concrete frame with masonry infill walls mid-rise               \\ 
\hline
C3H    & 6         & Nonductile reinforced concrete frame with masonry infill walls high-rise              \\ 
\hline
DS     & 1         & Rectangular cut-stone masonry block                                                   \\ 
\hline
INF    & 32        & Informal constructions.                                                               \\ 
\hline
M      & 23        & Mud walls                                                                             \\ 
\hline
RE     & 3         & Rammed Earth/Pneumatically impacted stabilized earth                                  \\ 
\hline
RM     & 2         & Reinforced masonry                                                                    \\ 
\hline
RS     & 21        & Rubble stone (field stone) masonry                                                    \\ 
\hline
RS1    & 3         & Local field stones dry stacked (no mortar) with timber floors, earth, or metal roof.  \\ 
\hline
RS2    & 1         & Local field stones with mud mortar.                                                   \\ 
\hline
RS3    & 3         & Local field stones with lime mortar.                                                  \\ 
\hline
S      & 9         & Steel                                                                                 \\ 
\hline
S1L    & 1         & Steel moment frame low-rise                                                           \\ 
\hline
S1M    & 1         & Steel moment frame mid-rise                                                           \\ 
\hline
S3     & 8         & Steel light frame                                                                     \\ 
\hline
S5     & 1         & Steel frame with unreinforced masonry infill walls                                    \\ 
\hline
UCB    & 39        & Concrete block unreinforced masonry with lime or cement mortar                        \\ 
\hline
UFB    & 33        & Unreinforced fired brick masonry                                                      \\ 
\hline
UFB1   & 1         & Unreinforced brick masonry in mud mortar without timber posts                         \\ 
\hline
W      & 28        & Wood                                                                                  \\ 
\hline
W1     & 2         & Wood stud-wall frame with plywood/gypsum board sheathing.                             \\ 
\hline
W2     & 1         & Wood frame, heavy members (with area  5000 sq. ft.)                                   \\ 
\hline
W3     & 5         & Wood light unbraced post and beam frame.                                              \\ 
\hline
W5     & 31        & Wattle and Daub (Walls with bamboo/light timber log/reed mesh and post).              \\
\hline
\end{tabular}
\end{table}

\newpage

\subsection{Completeness of Satellite Imagery}
\label{appendix_completness}

\begin{table}[!htpb]
\centering
\caption{Availability of Sentinel-1 GRD VV and VH signals for each country. As a reference, 0, 1, 2, and 3 mean 'none', 'available', 'uncovered', and 'only VV signal is available', respectively.}
\vspace{0.2cm}
\begin{tabular}{l|c|c|c|c|c|c|c|c} 
    \hline
    Country & 2016 & 2017 & 2018 & 2019 & 2020 & 2021 & 2022 & 2023 \\ \hline
    AFG     & 1    & 1    & 1    & 1    & 1    & 1    & 1    & 1    \\ \hline
    AGO     & 1    & 1    & 1    & 1    & 1    & 1    & 2    & 2    \\ \hline
    BDI     & 1    & 1    & 1    & 1    & 1    & 1    & 1    & 1    \\ \hline
    BEN     & 1    & 1    & 1    & 1    & 1    & 1    & 1    & 1    \\ \hline
    BFA     & 1    & 1    & 1    & 1    & 1    & 1    & 1    & 1    \\ \hline
    BGD     & 1    & 1    & 1    & 1    & 1    & 1    & 1    & 1    \\ \hline
    BTN     & 1    & 1    & 1    & 1    & 1    & 1    & 1    & 1    \\ \hline
    CAF     & 1    & 1    & 1    & 1    & 1    & 1    & 1    & 1    \\ \hline
    COD     & 0    & 1    & 1    & 1    & 1    & 1    & 1    & 1    \\ \hline
    COM     & 0    & 1    & 1    & 1    & 1    & 1    & 0    & 0    \\ \hline
    DJI     & 1    & 1    & 1    & 1    & 1    & 1    & 1    & 1    \\ \hline
    ERI     & 1    & 1    & 1    & 1    & 1    & 1    & 1    & 1    \\ \hline
    ETH     & 1    & 1    & 1    & 1    & 1    & 1    & 1    & 1    \\ \hline
    GIN     & 1    & 1    & 1    & 1    & 1    & 1    & 1    & 1    \\ \hline
    GMB     & 1    & 1    & 1    & 1    & 1    & 1    & 1    & 1    \\ \hline
    GNB     & 1    & 1    & 1    & 1    & 1    & 1    & 1    & 1    \\ \hline
    HTI     & 1    & 1    & 1    & 1    & 1    & 1    & 1    & 1    \\ \hline
    KHM     & 1    & 1    & 1    & 1    & 1    & 1    & 1    & 1    \\ \hline
    KIR     & 3    & 3    & 3    & 3    & 3    & 3    & 3    & 3    \\ \hline
    LAO     & 1    & 1    & 1    & 1    & 1    & 1    & 1    & 1    \\ \hline
    LBR     & 1    & 1    & 1    & 1    & 1    & 1    & 1    & 1    \\ \hline
    LSO     & 1    & 1    & 1    & 1    & 1    & 1    & 1    & 1    \\ \hline
    MDG     & 1    & 1    & 1    & 1    & 1    & 1    & 1    & 1    \\ \hline
    MLI     & 1    & 1    & 1    & 1    & 1    & 1    & 1    & 1    \\ \hline
    MMR     & 1    & 1    & 1    & 1    & 1    & 1    & 1    & 1    \\ \hline
    MOZ     & 1    & 1    & 1    & 1    & 1    & 1    & 1    & 1    \\ \hline
    MRT     & 1    & 1    & 1    & 1    & 1    & 1    & 1    & 1    \\ \hline
    MWI     & 1    & 1    & 1    & 1    & 1    & 1    & 1    & 1    \\ \hline
    NER     & 1    & 1    & 1    & 1    & 1    & 1    & 1    & 1    \\ \hline
    NPL     & 1    & 1    & 1    & 1    & 1    & 1    & 1    & 1    \\ \hline
    RWA     & 1    & 1    & 1    & 1    & 1    & 1    & 1    & 1    \\ \hline
    SDN     & 1    & 1    & 1    & 1    & 1    & 1    & 1    & 1    \\ \hline
    SEN     & 1    & 1    & 1    & 1    & 1    & 1    & 1    & 1    \\ \hline
    SLB     & 1    & 1    & 1    & 1    & 1    & 1    & 0    & 0    \\ \hline
    SLE     & 1    & 1    & 1    & 1    & 1    & 1    & 1    & 1    \\ \hline
    SOM     & 1    & 1    & 1    & 1    & 1    & 1    & 1    & 1    \\ \hline
    SSD     & 1    & 1    & 1    & 1    & 1    & 1    & 1    & 1    \\ \hline
    STP     & 1    & 1    & 1    & 1    & 1    & 1    & 1    & 1    \\ \hline
    TCD     & 1    & 1    & 1    & 1    & 1    & 1    & 1    & 1    \\ \hline
    TGO     & 1    & 1    & 1    & 1    & 1    & 1    & 1    & 1    \\ \hline
    TLS     & 1    & 1    & 1    & 1    & 1    & 1    & 1    & 1    \\ \hline
    TUV     & 3    & 3    & 3    & 3    & 3    & 3    & 3    & 3    \\ \hline
    TZA     & 1    & 1    & 1    & 1    & 1    & 1    & 1    & 1    \\ \hline
    UGA     & 1    & 1    & 1    & 1    & 1    & 1    & 1    & 1    \\ \hline
    VUT     & 1    & 1    & 1    & 1    & 1    & 1    & 1    & 1    \\ \hline
    YEM     & 1    & 1    & 1    & 1    & 1    & 1    & 1    & 1    \\ \hline
    ZMB     & 1    & 1    & 1    & 1    & 1    & 1    & 1    & 1    \\ \hline
\end{tabular}
\end{table}

\end{document}